\documentclass{article}

\PassOptionsToPackage{numbers, compress}{natbib}

\usepackage[preprint]{neurips_2026}

\usepackage[utf8]{inputenc}
\usepackage[T1]{fontenc}   
\usepackage{hyperref}      
\usepackage{url}           
\usepackage{booktabs}      
\usepackage{amsfonts}      
\usepackage{nicefrac}      
\usepackage{microtype}     
\usepackage{xcolor}        

\usepackage{amsmath}      
\usepackage{algorithm}     
\usepackage{algorithmicx}  
\usepackage{algpseudocode} 
\usepackage{dsfont}
\usepackage{graphicx}      
\usepackage{siunitx}

\newcommand{\classifier}{\varphi}

\newcommand{\data}{\mathcal{D}}
\newcommand{\target}{\mathcal{T}}
\newcommand{\shared}{\mathcal{R}}
\newcommand{\train}{\mathcal{X}}
\newcommand{\synthetic}{\mathcal{S}}

\title{ReMIA: a Powerful and Efficient Alternative to Membership Inference Attacks against Synthetic Data Generators}

\author{%
  Davide Scassola \\
  Aindo SpA \\
  Trieste, Italy \\
  \texttt{davide.scassola1@gmail.com} \\
  \And
  Andrea Coser \\
  Aindo SpA \\
  Trieste, Italy \\
  \texttt{andrea@aindo.com} \\
  \And
  Sebastiano Saccani \\
  Aindo SpA \\
  Trieste, Italy \\
  \texttt{sebastiano@aindo.com} \\
}

\begin{document}

\maketitle

\begin{abstract}
  Tabular data sharing under privacy constraints is increasingly important for research and collaboration. Synthetic data generators (SDGs) are a promising solution, but synthetic data remains vulnerable to attacks, such as membership inference attacks (MIAs), which aim to determine whether a specific record was part of the training data.
State-of-the-art MIAs are powerful but impractical: they rely on shadow modeling, requiring hundreds of SDG training runs, and need auxiliary data several times larger than the original training set.
Fast proxy metrics like distance to closest record (DCR) are efficient but have limited sensitivity to MIA risk.
We introduce ReMIA (Relative Membership Inference Attack), a practical privacy metric that requires only two SDG training runs and additional data no larger than the original training set.
Rather than predicting whether a record was in the training set, ReMIA generates two synthetic datasets from two source datasets and measures whether a classifier can identify which source a record came from. Experiments across multiple tabular datasets and SDGs show that ReMIA has a sensitivity comparable to state-of-the-art MIAs while being substantially more practical.
We further observe that SDGs can achieve privacy-utility trade-offs that traditional noise-based anonymization methods do not match.
Code is available at \url{https://github.com/aindo-com/remia}.

\end{abstract}

\section{Introduction}
Tabular data is ubiquitous across many fields and plays a fundamental role in scientific discovery and data-driven decision-making.
However, privacy concerns and regulations such as the European Union's General Data Protection Regulation (GDPR) limit the sharing and use of such data.
Synthetic data generation (SDG) \citep{jordon2022synthetic, hernandez2022synthetic}, often based on generative models \citep{xu2019modeling}, has emerged as a promising privacy-enhancing technology that allows for the creation and sharing of synthetic datasets that mimic the statistical properties of real data while protecting individual privacy \citep{bellovin2019privacy}.
Nevertheless, synthetic data is not intrinsically immune to privacy risks, as SDGs may leak information about individual records in the training data \citep{carlini2019secret, stadler2022synthetic}. 
Hence, being able to evaluate the privacy risk of synthetic data is crucial for its safe deployment \citep{giomi2023unified}, although this has proven challenging \citep{yao2025dcr, stadler2022synthetic}.

Privacy risk of synthetic data can be empirically assessed by evaluating its vulnerability to adversarial attacks, especially membership inference attacks (MIAs) \citep{homer2008resolving, shokri2017membership, salem2018ml, stadler2022synthetic, hu2022membership} where an attacker tries to infer whether a given record was part of the training data used by the model to generate the synthetic data.
MIAs are a powerful and principled way to assess privacy risk, but they are often computationally intensive and data-hungry.
In particular, shadow-modeling MIAs require hundreds of SDG training runs and large auxiliary datasets \citep{stadler2022synthetic, houssiau2022tapas, meeus2023achilles}. Alternatives have been proposed, such as DOMIAS \citep{van2023membership} that does not assume access to the generator, but still relies on auxiliary data and potentially expensive density estimation. On the other hand, fast proxy metrics such as distance to the closest record (DCR) \citep{park2018data, kotelnikov2023tabddpm, platzer2021holdout} are computationally efficient but have limited sensitivity to MIA risk \citep{yao2025dcr}.
Therefore, there is a need for a practical privacy metric that retains MIA-like sensitivity while significantly reducing computational and data requirements. 

We propose a new privacy risk metric, ReMIA (Relative Membership Inference Attack), inspired by MIAs but based on a different attack model: given two synthetic datasets produced by the same SDG from two different private training subsets, the goal of ReMIA is to correctly infer if a given record belongs to the training dataset relative to the first or the second synthetic dataset.
A high success rate of ReMIA implies that the synthetic datasets, and hence the SDG, are leaking information about the training data.
ReMIA requires training and generating with the SDG only twice, and the amount of additional data required is at most the size of the training dataset.
In particular, for a dataset with $n$ records, ReMIA can evaluate privacy while training each SDG instance on at least $\frac{n}{2}$ records, unlike many existing MIAs which require a large auxiliary dataset, meaning only a much smaller fraction of the available data is left for training the generator.
This makes ReMIA suitable for realistic settings where data is limited and generator training is costly.

We evaluate ReMIA across multiple tabular datasets, SDGs, and privacy metrics. Results show that ReMIA's privacy score is a proxy for the performance of state-of-the-art MIAs, while being substantially more practical to compute.
As a further empirical finding, we observe that modern SDGs can achieve better privacy-utility trade-offs than traditional noise-based anonymization techniques.
We summarize our contributions as follows:
\begin{itemize}
  \item We propose ReMIA, a practical metric for measuring privacy risk of synthetic data generators, inspired by MIAs but based on a novel attack model, requiring only two synthetic dataset generations and at most as much additional data as the training set.
  \item We provide an extensive empirical evaluation across several datasets, synthetic data generators and privacy metrics, showing that ReMIA correlates with strong MIA baselines while being significantly more efficient in computation and data usage.
  \item We show that, according to different privacy metrics, synthetic data can offer a better trade-off between privacy and realism than traditional noise-based anonymization methods.
\end{itemize}

\section{Background}

\subsection{Synthetic Data Generation}
Let us consider a dataset $\data$ containing sensitive information where records $x_i$ are samples from an unknown distribution $p(x)$.
Synthetic data generation is a privacy-enhancing technology (PET) that produces artificial records to share in place of sensitive data. Formally, a synthetic data generator (SDG) is a randomized function $\Phi$ that learns to approximate the underlying distribution that generated $\data$ and samples a new dataset $\data_s = \Phi(\data)$ with similar statistical properties but without revealing the original data.
The goal of an SDG is to produce synthetic data that is as useful as the original while preserving privacy.
Various approaches have been proposed for learning generative models for tabular data, such as statistical methods \citep{patki2016synthetic, zhang2017privbayes}, machine learning techniques \citep{nowok2016synthpop, watson2023adversarial}, and deep generative models \citep{kotelnikov2023tabddpm, xu2019modeling, lee2022differentially}.

\subsection{Privacy: Definition and Measurement}
The concept of privacy in data publishing spans legal, technical, and ethical dimensions. Under the GDPR (General Data Protection Regulation), data is considered anonymous only when it no longer carries the risk of identifying a person. More formally, information about individual records is leaked when it is possible to either (i) isolate a private record in the dataset, (ii) link two or more records concerning the same data subject, or (iii) deduce a sensitive attribute value from a set of other attributes.

In practice, there are two approaches to guarantee the safety of synthetic data.
One approach consists of designing SDGs that provide formal guarantees of protection, as those following the Differential Privacy (DP) \citep{dwork2006differential} framework, where calibrated noise is added to the private data used by the SDG. However, DP strongly limits the capability of SDGs in terms of fidelity of generated data \citep{adams2025fidelity, stadler2022synthetic}.

The alternative approach consists of empirically assessing the vulnerability of synthetic data a posteriori.
Methods based on the Distance to Closest Record (DCR) fall into this category \citep{park2018data, kotelnikov2023tabddpm, platzer2021holdout}. These assess leakage by measuring the similarity between the private training data and the synthetic data. Different definitions have been formulated in the literature, commonly based on measuring whether the DCR of training data with respect to the synthetic data is statistically smaller than when compared against a hold-out set of unseen data.
A common implementation measures the DCR of each record in the training data with respect to the synthetic data, and compares it with the DCR of records in a hold-out set of unseen data of the same size.
If the DCR to the synthetic data is smaller in significantly more than 50\% of comparisons, then the synthetic data is considered to be leaking information about the training data.
Despite being popular and computationally efficient, DCR-like metrics have shown limitations in detecting privacy leaks \citep{yao2025dcr}.
A more robust empirical assessment consists in measuring the effectiveness of adversarial attacks. In particular, we will focus on membership inference attacks (MIAs).

\subsection{Membership Inference Attacks}
Membership inference attacks (MIAs) \citep{homer2008resolving, shokri2017membership, salem2018ml, stadler2022synthetic, hu2022membership} are a class of linkage attacks where the adversary aims to determine whether a specific record $x$ was part of the private training dataset $\data$ used to train the model $\Phi$ that generated the synthetic data $\data_s = \Phi(\data)$.
In the common attack setting, the adversary has access to a target record $x \sim p(x)$ that may or may not be part of the training data $\data$ (in simulated attacks $P(x \in \data) = \frac{1}{2}$ is often assumed), the synthetic data $\data_s$, auxiliary data $\data_{\text{aux}}$ sampled from the same distribution as $\data$, and the (untrained) model $\Phi$ \citep{stadler2022synthetic}.
The vulnerability of a SDG to MIAs is then measured by simulating the attack multiple times and computing the attack's success rate, i.e., the fraction of times the adversary correctly identifies whether $x$ was in $\data$ or not.

MIAs can be implemented in various ways. Current state-of-the-art techniques against synthetic data generators are based on shadow modeling \citep{shokri2017membership}.
In MIAs based on shadow modeling (smMIAs), the adversary creates multiple shadow datasets $\data_{\text{shadow}}^i$ by sampling $n$ records from auxiliary data $\data_{\text{aux}}$, inserting the target record $x$ in half of them (in place of a random record). Each shadow dataset is used to train $\Phi$ and generate synthetic data $\data_s^i$, forming tuples $(\data_s^i, y^i)$ where $y^i$ indicates whether $x$ was included. A classifier $\phi$ is then trained to predict $y^i$ from $\data_s^i$, and finally applied to $\data_s = \Phi(\data)$ to estimate the probability that $x$ was part of the training data. Attacks differ in the choice of $\phi$ and feature extraction strategy.

\section{Related Work}
Literature on MIAs initially focused on discriminative models \citep{shokri2017membership} but only recently started to be applied to synthetic data generators \citep{stadler2022synthetic, hu2022membership}.
Early methods approximated the generator's output density to detect overfitting. \citet{hilprecht2019monte} proposed a Monte Carlo attack counting generated points near a target record, while \citet{chen2020gan} introduced GAN-Leaks, using the distance to the k-nearest synthetic neighbors as a density surrogate. \citet{hayes2019logan} introduced LOGAN, leveraging the generator's discriminator or a simple classifier to score records.

smMIAs were first introduced in \citet{shokri2017membership} as a powerful attack against machine learning models, and later adapted to synthetic data generators in \citet{stadler2022synthetic}, where the authors show that synthetic data generators are vulnerable to MIAs and that the success rate is correlated with the quality of the synthetic data. In this work, feature extraction is based on summary statistics, correlations, and marginal histograms of shadow synthetic datasets.
This framework was improved in \citet{houssiau2022tapas}, where the extracted features are the results of counting queries on the shadow synthetic datasets. In particular, the authors use k-way marginal statistics computed over subsets of the attribute values of the targeted record.
Building on this, \citet{meeus2023achilles} show how the success rate of smMIAs can be significantly improved by selecting outliers as target records.
Furthermore, \citet{guepin2023synthetic} relaxed the common requirement of a reference dataset, at the cost of a significant decrease in the attack's efficiency and efficacy.

In \citet{van2023membership}, the authors develop DOMIAS, a density-based MIA that exploits the local overfitting of the generative model to infer membership. DOMIAS also requires access to an auxiliary dataset, but not to the generative model, and it has been shown to be more effective than many MIA methods, while being more computationally efficient than smMIAs. Despite being an efficient way to measure privacy risk, DOMIAS has not been directly compared to smMIAs.

\section{Methodology}
\subsection{How Efficient are Privacy Metrics?}
MIAs based on shadow modeling (smMIAs) \citep{stadler2022synthetic, houssiau2022tapas, meeus2023achilles} are the state-of-the-art and hence the most sensitive metric to privacy leaks in generative models. However, they are prohibitively expensive: they require training hundreds of generative models on different datasets, along with a large auxiliary dataset that often exceeds the private training data, making them impractical as privacy metrics in real-world scenarios. Furthermore, these attacks target a single record at a time, making batch evaluation even costlier. For instance, \citet{meeus2023achilles} uses datasets of size 1,000 but requires 75,000 records in total to evaluate the attack.
DOMIAS \citep{van2023membership} is more practical, requiring only a single synthetic dataset, but still needs an auxiliary dataset and involves training two density models, which can be expensive unless the density model is very simple (for instance, a kernel density estimator). For instance, in one of their experiments training and control sets each contain 500 records, and the auxiliary set has 10,000 records, implying the training set is 22 times smaller than the total amount of data used.
DCR-based metrics are instead more efficient, requiring no additional SDG calls and only a control dataset of the same size as the training data.
Despite the recent advances in MIA research, there remains a need for a practical privacy metric that avoids expensive model training and does not require auxiliary data far exceeding the training set.

\subsection{ReMIA}
We aim to design a practical MIA-based metric that does not suffer from the aforementioned limitations.
To do this, we draw inspiration from the idea of MIAs and we propose a new type of attack: we evaluate the privacy of a generative method by testing whether an attacker can determine if a specific record belongs to one of two distinct training datasets from which two synthetic datasets were generated by the generative model.
If the attacker can successfully determine the origin of the record with high accuracy, it indicates that the generator is not privacy-preserving, as the synthetic data leaks specific information about individual records in the training data.
This is in contrast with MIAs, which test whether an attacker can determine if a specific record was part of the training data or not.
This approach allows us to evaluate the privacy of the generative method without requiring an auxiliary dataset, and without training the generative model more than twice.

We call our method ReMIA, which stands for "Relative Membership Inference Attack", since it is based on the idea of identifying which training dataset a record belongs to, rather than guessing whether a record was part of the training data or not as in standard MIAs.
Similar to MIAs, we can formalize an attack as a privacy game between a challenger $C$, having access to a private dataset $\data$ and a generative model $\Phi$, and an attacker $A$:
\begin{enumerate}
    \item $C$ builds two training datasets $\train_1 = (\target_1, \shared)$ and $\train_2 = (\target_2, \shared)$, where $\target_1$, $\target_2$, and $\shared$ are disjoint sets of records sampled from the same private dataset $\data$, with $|\target_1| = |\target_2|$.
    \item $C$ trains $\Phi$ on $\train_1$ and $\train_2$ to generate two synthetic datasets $\synthetic_1 = \Phi(\train_1)$ and $\synthetic_2 = \Phi(\train_2)$.
    \item $C$ sends $\synthetic_1$, $\synthetic_2$ and $\target = (\target_1, \target_2)$ (shuffled, without label information) to the attacker $A$.
    \item $A$ uses $\synthetic_1$ and $\synthetic_2$ to build a classifier $\classifier$ that predicts whether each record in $\target$ belongs to $\target_1$ or $\target_2$. The score of the attack is the performance of the classifier $\classifier$ on $\target$.
\end{enumerate}
If the attacker can successfully determine the origin of the record with better-than-random accuracy, it indicates that the generator is not privacy-preserving, as the synthetic data leaks specific information about individual records in the training data.
On the other hand, if the attacker cannot distinguish between $\synthetic_1$ and $\synthetic_2$ with better than random guessing, it suggests that the generator is effectively preserving privacy by generating synthetic data that does not depend on specific records in the training data.
Throughout the paper, we use the AUROC of the classifier $\classifier$ as the privacy score of ReMIA.

\subsection{An Attacker Model for ReMIA}
Building an attacker model for ReMIA is not straightforward, as we do not assume the attacker to have access to auxiliary data or the generative model.
We propose a simple method consisting in training a classifier $\classifier$ to distinguish between the two synthetic datasets $\synthetic_1$ and $\synthetic_2$, and then evaluating the performance of this classifier on the two unseen sets of records $\target_1$ and $\target_2$.

In practice, in order to observe a signal on unseen data $\target_1$ and $\target_2$, we train the classifier $\classifier$ past the point of overfitting on the synthetic data $\synthetic_1$ and $\synthetic_2$, and we stop when a certain threshold of performance on the training synthetic data is reached. For instance, we stop when the AUROC reaches 99\%.
Notice that this is not a problem, since we are not interested in the performance of the classifier on the synthetic data, but rather in its performance on the unseen data $\target_1$ and $\target_2$ as a "side effect".
The intuition behind this attack method is that if the classifier can successfully distinguish between $\synthetic_1$ and $\synthetic_2$, it may be able to learn features that are specific to the training datasets $\train_1$ and $\train_2$, which in turn may allow it to correctly classify records in $\target_1$ and $\target_2$ based on the synthetic data.
In order to reduce the noise of the privacy score, we apply a smoothing function to the performance of the classifier on $\target_1$ and $\target_2$ across the learning iterations, and select as the score the smoothed value at the iteration where the performance on the synthetic data reaches a certain threshold.

Notice that without having access to the labels of $\target$, the attacker cannot know at which training iteration the classifier reaches the best performance on the unseen data $\target$, as we show in the example in Figure \ref{remia_score_learning_step} in Appendix \ref{app:additional_results}. The stopping criterion based on the performance on the synthetic data is a practical heuristic that we will show to be effective in practice. In principle, one can select the maximum performance on the unseen data $\target$ as the privacy score. However, this can lead to a biased estimate of the privacy risk. This can be solved by assuming the attacker to have access to a subset of labeled records from $\target$ that can be used as a validation set to select the best iteration. We leave this analysis for future work.

We implement the classifier $\classifier$ as a multi-layer perceptron (MLP). More details about the architecture, hyperparameters, smoothing function, and the training procedure are provided in the experimental section and Appendix \ref{app:remia_details}.

\subsection{Efficiency of ReMIA}
Notice that ReMIA only requires training the generative model twice (on $\train_1$ and $\train_2$), it does not require any auxiliary dataset, but only a hold-out split of the original dataset, and the attacker model only consists in training a single classifier.
The data usage of ReMIA is also efficient. Let $s = |\train_1| = |\train_2|$ be the size of each training set, where $t = |\target_1| = |\target_2|$ is the number of target records in each training set, and $r = |\shared|$ is the number of shared records so that $s = t + r$. If the fraction of target records in each training dataset is $f = \frac{t}{s}$, then the total number of records needed to run ReMIA is $n = |\target_1|+|\target_2|+|\shared| = s(1+f)$, which is at most twice the size of the training dataset used to train the generative model.
This allows us to evaluate the privacy of the generative model on training datasets whose size is comparable to the original dataset. 
We can choose a fraction of target records $f$ that is large enough to obtain a statistically stable result, i.e., that does not depend on the specific set of target records. On the other hand, a small fraction allows us to evaluate ReMIA on a training dataset that is closer in size to the total amount of available data.

While smMIAs represent powerful but prohibitive privacy metrics, ReMIA achieves practicality by trading off the attack realism for computational efficiency.
We summarize the ReMIA privacy score computation in Algorithm \ref{alg:remia}.
\begin{algorithm}[t]
\caption{ReMIA privacy score computation}
\label{alg:remia}
\begin{algorithmic}[1]
\Require Dataset $\data$, generative model $\Phi$, targets fraction $f$, iterations $N$
\Ensure ReMIA privacy score
\State Randomly Split $\data$ into $\target_1$, $\target_2$, and $\shared$, with $|\target_1| = |\target_2| = \frac{f}{1+f}|\data|$ and $|\shared| = |\data| - 2|\target_1|$
\State Create two training datasets $\train_1 = \target_1 \cup \shared$ and $\train_2 = \target_2 \cup \shared$
\State Generate synthetic datasets $\synthetic_1 = \Phi(\train_1)$ and $\synthetic_2 = \Phi(\train_2)$
\State Create a dataset $\synthetic = \synthetic_1 \cup \synthetic_2$ and $\target = \target_1 \cup \target_2$ with origin labels for each record
\State Initialize a classifier $\classifier$
\For{$i = 1$ \textbf{to} $N$}
  \State Train $\classifier$ on $\synthetic$ for one iteration
  \State Evaluate AUROC of $\classifier$ on $\target$ to obtain $P_{\text{target}}^i$ and on $\synthetic$ to obtain $P_{\text{train}}^i$
\EndFor
\State Apply a smoothing function to scores $\{P^i_{\text{target}}\}_{i=1}^N$ to obtain smoothed scores $\{\tilde{P}^i_{\text{target}}\}_{i=1}^N$
\State \Return $\tilde P^k_{\text{target}}$ where $k = \min\{ i \mid P_{\text{train}}^i \geq 99\%\}$
\end{algorithmic}
\end{algorithm}

\section{Experiments}

\subsection{Experimental Setup}

\paragraph{Datasets.}
We evaluate the privacy metrics on three datasets previously used in MIAs literature: Adult \citep{adult} (5 numerical and 10 categorical columns), UK Census \citep{uk_census} (17 categorical columns), and California \citep{california} (9 numerical columns).

\paragraph{Generators.}
We evaluate the privacy metrics on several open-source synthetic data generators (SDGs), implemented in the \texttt{reprosyn} \citep{reprosyn2022} and \texttt{synthcity} \citep{synthcity} libraries: SynthPop \citep{nowok2016synthpop} based on AR models, BayNet \citep{zhang2017privbayes} based on Bayesian networks, PrivBayes \citep{zhang2017privbayes} which is a differentially private version of the BayNet, CTGAN \citep{xu2019modeling}, ADSGAN \citep{yoon2020anonymization} and PateGAN \citep{jordonpate} based on GANs \citep{goodfellow2020generative}, TVAE \citep{xu2019modeling} based on VAEs \citep{Kingma2014}, TabDDPM \citep{kotelnikov2023tabddpm} that uses a diffusion model \citep{ho2020denoising} and ARF \citep{watson2023adversarial} based on a form of unsupervised random forests \citep{ho1995random}.
Moreover, we also evaluate the performance of our proprietary generator \textit{Aindo}.

\paragraph{Risk models.}
We follow the idea of \textit{risk models} introduced in \citet{palacios2025empirical}, which consist of methods to deliberately introduce privacy risks in synthetic datasets with the purpose of assessing privacy quantification methods.
First, we use the \textit{leaky risk model} \citep{palacios2025empirical}, that generates a synthetic dataset by randomly selecting a fraction $p$ of records from the training set and the remaining records from a disjoint dataset of samples from the same distribution.
Secondly, as we are interested in evaluating the performance of noise-based anonymization techniques, we use the \texttt{aindo-anonymize} library \citep{aindo-anonymize}, which introduces independent noise to each feature of a row in a way that the marginal distributions of the original data are preserved. The level of noise is controlled by a parameter $\alpha \in [0,1]$.
More details are provided in Appendix \ref{app:risk_models}.

\paragraph{Privacy Evaluation Baselines.}
We compare our proposed metric against the previously introduced privacy metrics: DOMIAS and smMIAs.
For smMIA, we use the implementation of \citep{meeus2023achilles}. In particular, we use the query-based attack \citep{houssiau2022tapas} combined with a random forest classifier.
Moreover, we select as target record the "most outlier" record, as done in \citep{meeus2023achilles}, which is the record with the highest average cosine distance $d$ from the k-closest neighbors. This setting represents the state of the art for smMIAs \citep{meeus2023achilles}.
To observe the effectiveness of the attack against more typical targets, we also evaluate the attack on a target record whose $d$ is the median. We refer to these two attack settings as "smMIA(out)" and "smMIA(med)" respectively.
We set the size of the auxiliary dataset to be 10 to 50 times larger than the training dataset depending on data availability.

For DOMIAS, we use the original implementation \citep{van2023membership}. In particular, we use kernel density estimation (KDE) as density estimation method. Despite the authors showing the application of the flow-based BNAF \citep{de2020block} density estimator, it is considerably slower to train and it has been shown to be marginally more effective than KDE only with large datasets.
In order to apply KDE, we use one-hot encoding for categorical features and we eventually apply PCA to avoid dealing with singular matrices. We set the size of the auxiliary dataset to be 5 times larger than the training dataset.

In order to improve the comparability of the results, we fix the size of the training datasets (with which the generative models are trained) to 1000 for every privacy metric, and we set the size of the synthetic datasets to be the same as the original one, as is common practice. This is the same setting used in \citet{meeus2023achilles}.

\paragraph{Quality Evaluation.}
In order to evaluate the quality of the synthetic data, we use the following common metrics: (i) \textbf{Detection} which is the performance of a machine learning model trained to distinguish between real and synthetic data, and (ii) \textbf{ML efficacy} which is the performance of a machine learning model trained on the synthetic data and tested on the real data. In both cases we use an XGBoost model \citep{chen2015xgboost} as it is a widely used and effective model for tabular data.
For Detection, the lower the performance of the model, the better the fidelity of the synthetic data, as it means that the model is not able to distinguish between real and synthetic data. For ML efficacy, the higher the performance of the model, the better the utility of the synthetic data, as it means that the model is able to learn from the synthetic data and generalize to the real data.

More details on the experimental setup are provided in Appendix \ref{app:experimental_setup}.

\subsection{Results}

We evaluate the privacy risk of the SDGs by measuring the performance of the smMIA attacks, DOMIAS, DCR and ReMIA in terms of area under the ROC curve (AUROC), meaning a lower score indicates better privacy.
For computational reasons, we are not able to evaluate the smMIA attack on all the SDGs, as a single attack requires training a generative model 1200 times. We then show only the results of the smMIA attacks on a subset of the aforementioned SDGs (7 out of 11).
Unless otherwise specified, for ReMIA we use a fraction of target records of 1, but we also experiment with a fraction of 0.5. When necessary, we refer to these two settings as ReMIA($f=1.0$) and ReMIA($f=0.5$).
When possible, we repeat experiments for 4 different random seeds that affect the data splitting, training and sampling of SDGs, and privacy metrics computation.

\paragraph{Privacy Scores Comparison.}
We show in Table \ref{tab:privacy_scores_adult} the privacy scores obtained by each generator on the Adult dataset in terms of area under the ROC curve (AUROC).
The results for the other datasets are reported in Appendix \ref{app:additional_results}.
\begin{table}[t]
  \caption{Privacy scores (AUROC) for Adult dataset (\%). We show the mean and standard deviation of the scores across all repetitions. When the relative accuracy is significantly higher than 50\% according to a one-sided binomial test with $p<0.001$ the scores are underlined (we aggregate counts over all repetitions). Because of the computational cost, only one repetition was performed for \textit{Aindo}.}
  \label{tab:privacy_scores_adult}
  \centering
  \begin{tabular}{lS[table-format=2.1(1)]S[table-format=2.1(1)]S[table-format=2.1(1)]S[table-format=2.1(1)]S[table-format=2.1(1)]}
    \toprule
    Generator & \multicolumn{1}{c}{DCR} & \multicolumn{1}{c}{DOMIAS} & \multicolumn{1}{c}{ReMIA} & \multicolumn{1}{c}{smMIA(out)} & \multicolumn{1}{c}{smMIA(med)} \\
    \midrule
    ADSGAN & \num{14.6 +- 5.5} & \num{51.7 +- 0.8} & \underline{\num{54.8 +- 2.0}} & \multicolumn{1}{c}{-} & \multicolumn{1}{c}{-} \\
    ARF & \num{19.7 +- 0.4} & \underline{\num{53.3 +- 0.7}} & \underline{\num{59.2 +- 1.6}} & \underline{\num{62.6 +- 2.1}} & \num{51.2 +- 3.9} \\
    Aindo & \num{47.8 +- 2.3} & \underline{\num{54.2 +- 0.9}} & \underline{\num{60.1 +- 1.1}} & \num{63.3} & \num{54.4} \\
    BayNet & \underline{\num{76.4 +- 2.1}} & \underline{\num{78.4 +- 1.0}} & \underline{\num{79.4 +- 1.0}} & \underline{\num{82.8 +- 2.3}} & \underline{\num{82.0 +- 4.7}} \\
    CTGAN & \num{12.5 +- 1.1} & \num{51.0 +- 1.2} & \underline{\num{53.3 +- 1.0}} & \underline{\num{59.2 +- 2.5}} & \num{54.2 +- 4.2} \\
    PateGAN & \num{15.1 +- 10.6} & \num{50.7 +- 0.6} & \num{51.7 +- 2.6} & \multicolumn{1}{c}{-} & \multicolumn{1}{c}{-} \\
    PrivBayes($\epsilon$=10) & \num{0.7 +- 0.3} & \num{50.4 +- 2.6} & \num{51.0 +- 1.5} & \num{51.1 +- 2.5} & \num{50.4 +- 4.3} \\
    PrivBayes($\epsilon$=1000) & \num{34.0 +- 1.5} & \underline{\num{59.5 +- 1.3}} & \underline{\num{60.7 +- 0.9}} & \underline{\num{80.3 +- 1.8}} & \underline{\num{66.0 +- 4.2}} \\
    SynthPop & \underline{\num{56.6 +- 1.9}} & \underline{\num{59.5 +- 2.5}} & \underline{\num{67.4 +- 0.6}} & \underline{\num{79.0 +- 4.7}} & \underline{\num{66.9 +- 3.7}} \\
    TVAE & \num{13.0 +- 2.3} & \num{51.0 +- 1.6} & \underline{\num{54.2 +- 0.4}} & \multicolumn{1}{c}{-} & \multicolumn{1}{c}{-} \\
    TabDDPM & \num{0.5 +- 0.3} & \num{50.0 +- 0.3} & \num{50.3 +- 1.1} & \multicolumn{1}{c}{-} & \multicolumn{1}{c}{-} \\
    \bottomrule
\end{tabular}

\end{table}
We first observe that smMIA(out) is able to detect a significant privacy risk in almost all the generators, and its score is almost always the highest. This is in line with \citet{meeus2023achilles}, which shows that smMIA is a powerful attack. However, smMIA(med) has a significantly weaker signal and is not able to detect a significant privacy risk in many generators.
ReMIA and DOMIAS also detect a privacy risk in many generators and they appear to be comparable with smMIA(med), with ReMIA having a slightly stronger signal.
Indeed, ReMIA has a significant (Spearman) correlation with DOMIAS and with the smMIA(med) attack as we show in Table \ref{privacy_scores_correlation}.
\begin{table}[t]
  \caption{Spearman correlation between the scores of the different metrics across datasets and SDGs. The correlation is computed on the mean scores across all repetitions, stars indicate statistical significance (**: $p<0.01$, ***: $p<0.001$). Scores inferior to 50\% are clipped to 50\% before computing the correlation.}
  \label{privacy_scores_correlation}
  \centering
  \begin{tabular}{llllll}
    \toprule
    Metric & DOMIAS & ReMIA(f=1.0) & ReMIA(f=0.5) & smMIA(out) & smMIA(med) \\
    \midrule
    DOMIAS & - & 0.91*** & 0.86*** & 0.12 & 0.70*** \\
    ReMIA(f=1.0) & 0.91*** & - & 0.94*** & 0.17 & 0.64** \\
    ReMIA(f=0.5) & 0.86*** & 0.94*** & - & 0.12 & 0.57** \\
    smMIA(out) & 0.12 & 0.17 & 0.12 & - & 0.00 \\
    smMIA(med) & 0.70*** & 0.64** & 0.57** & 0.00 & - \\
    \bottomrule
\end{tabular}

\end{table}
Moreover, we observe that the score of ReMIA is highly correlated when using half of the target records (ReMIA($f=0.5$)), meaning that ReMIA allows us to reduce the amount of additional data required to obtain a reliable privacy risk estimation.

We summarize the power comparison of the metrics in Table \ref{tab:metrics_power_comparison_summary}, where we show that ReMIA is slightly more sensitive than DOMIAS, similarly sensitive to smMIA(med) but less sensitive than smMIA(out).
Moreover, our experiments confirm that DCR does not provide a reliable indication of the privacy risk of SDGs.
\begin{table}
  \caption{Metrics Sensitivity Comparison. The first column shows the number of dataset-generator pairs where the score was greater than 0.5 for every repetition.
  The second column shows mean and standard deviation of the difference between the score of each metric and the score of ReMIA(f=1.0) metric across all dataset-generator pairs.
  Scores inferior to 50\% are clipped to 50\%.}
  \label{tab:metrics_power_comparison_summary}
  \centering
  \begin{tabular}{lll}
    \toprule
     & Risk Detected (\%) & Score Difference vs ReMIA (\%) \\
    \midrule
    ReMIA(f=1.0) & 72.7\% (24/33) & - \\
    ReMIA(f=0.5) & 72.7\% (24/33) & \num{0.6 +- 1.9} \\
    DCR & 9.1\% (3/33) & \num{-4.7 +- 4.0} \\
    DOMIAS & 66.7\% (22/33) & \num{-2.1 +- 2.3} \\
    smMIA(out) & 90.5\% (19/21) & \num{16.0 +- 12.6} \\
    smMIA(med) & 42.9\% (9/21) & \num{-1.1 +- 7.6} \\
    \bottomrule
\end{tabular}

\end{table}

\begin{figure}[t]
  \centering
  \includegraphics[width=1.0\textwidth]{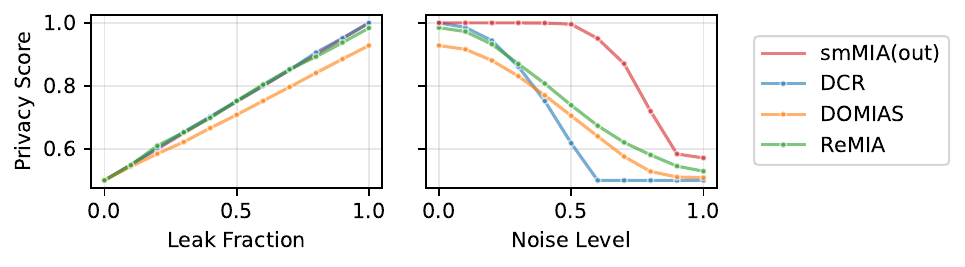}
  \caption{Metrics evaluation in risk models (Adult dataset).
  The plots show the privacy risk (AUROC score averaged over 4 seeds) of different metrics for different values of the leak fraction $p$ of the leaky model (left) and noise level $\alpha$ of the noise-based anonymization method (right). All metrics' scores are monotonic with respect to both $p$ and $\alpha$, they follow the maximum achievable by a MIA ($0.5 + \frac{p}{2}$) in the leaky model, while in the anonymized model smMIA(out) score is higher than ReMIA and DOMIAS, that are similar, while DCR performs poorly. Scores inferior to 50\% are clipped to 50\%. Results on the other datasets are reported in Appendix \ref{app:additional_results}.
  }
  \label{leak_fraction_alpha_vs_privacy_adult}
\end{figure}

Finally, we show in Figure \ref{leak_fraction_alpha_vs_privacy_adult} that the privacy risk detected by ReMIA is monotonic with respect to the leak fraction $p$ of the leaky model and the noise level of the noise-based anonymization method. The score of ReMIA in the leaky model closely follows $\frac{1+p}{2}$, the maximum achievable by a MIA.

\begin{figure}[t]
  \centering
  \includegraphics[width=1.0\textwidth]{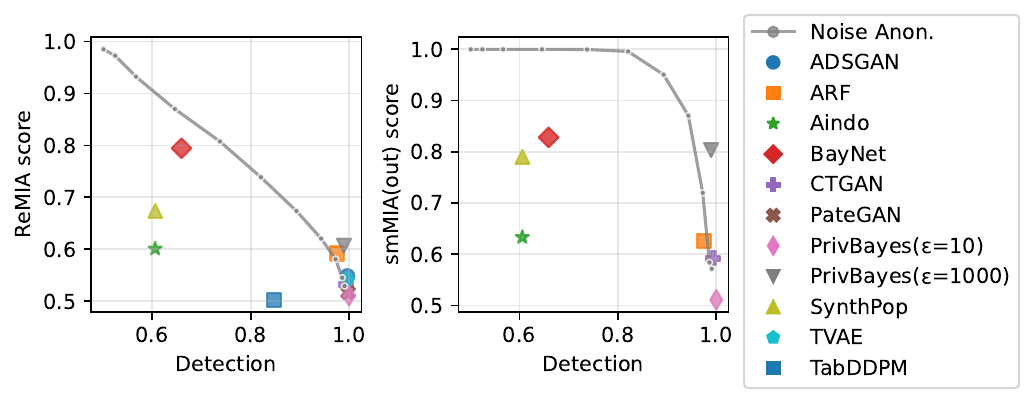}
  \caption{Privacy-fidelity trade-off (Adult dataset). We show the fidelity of the synthetic data in terms of Detection (lower is better) against the privacy scores of ReMIA (left) and smMIA(out) (right). The gray line is the performance of the noise-based anonymization method, meaning that models below the line achieve a better privacy-fidelity trade-off. Results on the other datasets and privacy metrics are reported in Appendix \ref{app:additional_results}.
  }
  \label{fidelity_vs_privacy_adult}
\end{figure}

\paragraph{Privacy-Quality Trade-off.}
We have seen how few models are able to achieve a low privacy risk, but all models that do not present a significant privacy risk generate poor quality synthetic data, both in terms of fidelity and utility. This trade-off between privacy and fidelity is a well-known phenomenon \citep{stadler2022synthetic}. However, there are some models that achieve a better trade-off than others, as we show in Figure \ref{fidelity_vs_privacy_adult}.
Furthermore, in contrast with the conclusions of \citet{stadler2022synthetic} and in line with those of \citet{van2023membership}, we observe that some SDGs are able to achieve a better trade-off than the noise-based anonymization technique, as they are able to achieve a low privacy risk while maintaining a good quality of the synthetic data. More experiments are reported in Appendix \ref{app:additional_results}.

\paragraph{Efficiency Comparison.}
ReMIA requires only a fraction of the computational time and data of smMIA attacks. Moreover, our method is also more data-efficient than DOMIAS, but it requires training the generative model two times. The overhead time due to training the discriminator is higher than DOMIAS with KDE, but still relatively low.
We show in Table \ref{tab:efficiency_comparison} the comparison of the efficiency of the metrics in terms of computational time, number of models trained and additional data required.
\begin{table}[t]
  \caption{Efficiency comparison. The additional data usage is computed as $\frac{n}{s} - 1$ where $n$ is the total data usage and $s$ is the size of the training datasets. The overhead time is the total time required for running the metric excluding the training of the SDGs. In both cases we report the range of values if they differ across experiments. In the case of smMIAs $n$ depends on the particular dataset. The wide range in overhead time also depends on the dataset used.}
  \label{tab:efficiency_comparison}
  \centering
  \begin{tabular}{llll}
    \toprule
    Metric & SDG Calls & Additional Data Usage & Overhead Time (s) \\
    \midrule
    DCR & 1 & 1.0 & <0.2 \\
    DOMIAS & 1 & 6.0 & 2.7 - 3.6 \\
    ReMIA(f=0.5) & 2 & 0.5 & 7.8 - 23.0 \\
    ReMIA(f=1.0) & 2 & 1.0 & 8.4 - 23.3 \\
    smMIA & 1200 & 19.6 - 74.0 & 16.4 - 4108.0 \\
    \bottomrule
\end{tabular}

\end{table}
Overall, ReMIA represents a practical alternative to MIAs that values computational and data efficiency while maintaining a high sensitivity to privacy risks.

\section{Discussion}
We propose ReMIA, a new privacy metric inspired by MIAs but based on a new attack model that requires significantly less computational and data resources than state-of-the-art MIAs, making it a practical tool for evaluating the privacy of synthetic data generators in realistic settings.
Across multiple datasets and generators, our experiments show that ReMIA correlates with state-of-the-art MIAs and achieves competitive sensitivity relative to lower-cost alternatives to MIAs based on shadow modeling.
Moreover, we find empirical evidence that synthetic data generated by SDGs can obtain a better privacy-quality trade-off than noise-based anonymized data according to different privacy metrics.

\paragraph{Limitations.}
Like most privacy metrics, ReMIA depends on implementation choices, including classifier design and training protocol.
Moreover, extending comparisons to a broader range of SDGs, anonymization methods, datasets, training sizes, and attack settings remains limited by the cost of running smMIAs and the cost of training powerful SDGs. In particular, a more extensive exploration of the effect of training size on privacy risk is an interesting direction for future work.

\paragraph{Future work.}
Future work should focus on studying the theoretical connections between ReMIA, overfitting, and MIAs, to better understand when ReMIA or detection of overfitting are good proxies for privacy risk.
On the methodological side, stronger attacker models may increase sensitivity while preserving computational efficiency.
Finally, the use of outliers as in \citet{meeus2023achilles} for improving the performance of ReMIA is an interesting direction to explore.

\paragraph{Broader impact.}
As a practical tool for evaluating the privacy of synthetic data generators, ReMIA can help researchers and practitioners to better detect and mitigate privacy risks associated with synthetic data. By providing a more computationally accessible metric for privacy evaluation, it may encourage the development and adoption of safer synthetic data generation techniques, ultimately contributing to the responsible use of synthetic data in various applications. On the other hand, like any empirical metric, it may not capture all aspects of privacy risk, and its misuse or misinterpretation could lead to a false sense of security.

\begin{ack}
This project was supported by the European Union through grant agreement 101218531 - SyDai.
\end{ack}

\bibliographystyle{plainnat}
\bibliography{references}

\newpage
\appendix
\section{Methods and Experimental Setup Details}
\label{app:experimental_setup}

\subsection{ReMIA Details}
\label{app:remia_details}
ReMIA trains a tabular discriminator to distinguish between synthetic data from two different sources, and it is then applied to a non-intersecting subset of the training records of each source to assess the information leakage of the synthetic data.
Inputs are preprocessed as follows: numerical features are standardized using training-set mean and standard deviation, and categorical features are one-hot encoded (unknown categories are mapped to zero vectors of the same dimension).
The discriminator is a multilayer perceptron with two hidden layers (100 and 50 units), SiLU activations \citep{elfwing2018sigmoid}, and layer normalization \citep{ba2016layer} applied before each hidden layer. The output layer has a single unit with sigmoid activation.

Optimization uses RAdam \citep{loshchilov2017decoupled} with learning rate $5\times 10^{-3}$ and binary cross-entropy. Training runs for at most $1000$ epochs, with batch size $500$ and early stopping on training loss with patience $50$.
The AUROC score is computed on the training set, and on the target records every 2 steps, using the discriminator's predicted probabilities as membership scores. In order to improve stability and reduce noise, AUROC scores are smoothed by a centered rolling mean with a window equal to $10\%$ of the recorded steps.

\subsection{Baseline Privacy Metrics}
\label{app:baseline_privacy_metrics}

\paragraph{DCR.}
Given a distance function $d$, a record $x$ and a dataset $\data$, we define the distance to closest record (DCR) as $\text{DCR}(x, \data) = \min_{x' \in \data} d(x, x')$.
In this article, we implement the DCR-based privacy metric following \citet{zhang2024mixed}:
Given training data $\data_t$, synthetic data $\data_s$, hold-out data $\data_h$, and a distance function $d$, we compute for each record $x_i$ in $\data_t$ the DCRs to the synthetic data $\text{DCR}(x_i, \data_s)$ and to the hold-out data $\text{DCR}(x_i, \data_h)$, and we compute the fraction of records in $\data_t$ for which the DCR to the synthetic data is smaller than the DCR to the hold-out data:
\begin{equation}
\frac{1}{|\data_t|} \sum_{x_i \in \data_t} \mathds{1}[\text{DCR}(x_i, \data_s) < \text{DCR}(x_i, \data_h)]
\end{equation}
If this fraction is significantly higher than 50\%, then the synthetic data is considered to be leaking information about the training data.
In our implementation, we use the cosine distance (1 minus the cosine similarity) as the distance function, after standardizing numerical features and one-hot encoding categorical features.

\paragraph{smMIA.}
We use the implementation of the MIA based on shadow modeling as described in \citet{meeus2023achilles}.
In particular, we use the query-based feature extraction \citep{houssiau2022tapas} and a random forest as the predictive model because it represents the most effective attack.
We keep the default settings where the number of training shadow datasets is set to 1000, and the number of test shadow datasets is set to 200. In half of both training and test shadow datasets, the target record is included in the training set of the shadow model, while in the other half it is not. The shadow models are trained for 100 epochs with early stopping on validation loss with patience 10, and the best model is selected based on validation performance.
Data for building the shadow datasets is sampled with replacement from the original dataset, in particular we use two different splits of the original dataset for building the training and test shadow datasets, where the split for the training data is double the size of the split for the test data. In particular, we use all the available data for the Adult and California datasets, while for the UK Census dataset we use a random sample of 75,000 records. Considering training shadow datasets with 1,000 records, this means that the total data usage for running the attack is $\approx20\times$ for California dataset, $\approx40\times$ for Adult dataset, and $\approx75\times$ for the UK Census dataset.

We follow the strategy outlined in \citep{meeus2023achilles} for the selection of the target record, which consists of sorting the records by the average cosine distance for the k-nearest neighbors and selecting the one with the highest distance.
This strategy is based on the intuition that records that are more isolated in the feature space are more vulnerable to MIAs, and thus represent a worst-case scenario for the evaluation of the privacy risk of synthetic data. Indeed, it has been shown that the performance of MIAs is significantly higher on outlier records than on random records \citep{meeus2023achilles}.
In order to evaluate the performance of smMIA on a more typical scenario, we also evaluate the attack on the record with the median distance to its k-nearest neighbors. This makes smMIA more comparable to the other metrics, which are not designed to target single outlier records.
Performing more attacks on different target records would be computationally prohibitive. In many cases, one smMIA takes several hours to be performed. As we later report in Appendix \ref{app:computational_resources}, the total time taken for the experiments shown in this article is approximately 882 hours, of which about 865 hours were spent on attacks based on shadow modeling.

\paragraph{DOMIAS.}
We adapt the original implementation of DOMIAS \citep{van2023membership} to tabular data with discrete features.
To do this, we first encode categorical features with one-hot encoding. Then, we apply principal component analysis (PCA) to avoid linearly dependent features that cause issues when training density estimators.
We use kernel density estimation (KDE) with Gaussian kernel as the density estimator, keeping the settings of the original implementation.
We do not use the block neural autoregressive flow (BNAF) \citep{de2020block} as the density estimator because it is designed for continuous data and from preliminary experiments it performed poorly compared to KDE. This was also observed in the original paper, where the advantage of using BNAF over KDE was emerging only for large training datasets,
making it not competitive considering the substantial increase in computational cost.
Finally, we set the size of the reference (auxiliary) dataset to be 5 times larger than the training set (so in our case 5000 records).
These settings imply a total additional data usage that is 6 times the size of the training set, considering the reference dataset ($5\times$) and the control dataset ($1\times$) containing non-training records used to evaluate the performance.

\subsection{Synthetic Data Generators}
\label{app:sdgs}
We show in Table \ref{tab:sdgs_summary} the list of SDGs used in our experiments, with a reference to their respective publications, the implementation used, and the maximum running time across all experiments (for datasets of size 1,000).
\begin{table}[h]
  \centering
  \caption{Synthetic data generators used in our experiments, with implementation sources and maximum observed runtimes for datasets of size 1,000.}
  \label{tab:sdgs_summary}
  \begin{tabular}{lccc}
    \toprule
    SDG Name & Reference & Implementation & Runtime (seconds)\\
    \midrule
    ADSGAN & \citet{yoon2020anonymization} & synthcity & 222.0 \\
    ARF & \citet{watson2023adversarial} & synthcity & \phantom{0}25.4 \\
    Aindo & - & proprietary & 153.9 \\
    BayNet & \citet{zhang2017privbayes} & rsyn & \phantom{0}14.5 \\
    CTGAN & \citet{xu2019modeling} & rsyn & \phantom{0}16.7 \\
    PateGAN & \citet{jordonpate} & synthcity & 494.2 \\
    PrivBayes($\varepsilon=10$) & \citet{zhang2017privbayes} & rsyn & \phantom{0}14.6 \\
    PrivBayes($\varepsilon=1000$) & \citet{zhang2017privbayes} & rsyn & \phantom{0}16.3 \\
    SynthPop & \citet{nowok2016synthpop} & rsyn & \phantom{00}6.2 \\
    TVAE & \citet{xu2019modeling} & synthcity & \phantom{0}86.0 \\
    TabDDPM & \citet{kotelnikov2023tabddpm} & synthcity & \phantom{0}84.0 \\
    \bottomrule
  \end{tabular}
\end{table}

\subsection{Privacy Risk Models}
\label{app:risk_models}

\paragraph{Leaky model.}
We use the leaky risk model as defined in \citet{palacios2025empirical}: given a training dataset $D_{t}$ and a disjoint dataset of samples from the same distribution $D_{c}$ of the same size $n$, the leaky risk model $L_p$ generates a synthetic dataset $D_{s} = L_f(D_{t})$ by randomly selecting $p \cdot n$ rows from $D_{t}$ and $(1-p) \cdot n$ rows from $D_{c}$.
This model was introduced to control the amount of privacy risk in the synthetic dataset, by leaking a certain amount of rows from the training set. When $p=0$ this model simulates the perfect synthetic data generator, while when $p=1$ the training data is completely leaked, that is the worst possible scenario.
Notice that for all $p \in [0,1]$ the quality of the "generated" data is maximal.

It is possible to compute the theoretical maximum accuracy of a MIA against this model, which is $\frac{1 + p}{2}$.
In this case, the best possible attack strategy is to predict membership (1) when an exact match of the target record is found in the synthetic dataset, and to predict non-membership (0) otherwise. The only case where this strategy fails is when the target record is in the training set but it is not leaked for the synthetic dataset, which happens with probability $\frac{1}{2}(1-p)$,
meaning that the attack succeeds with probability $1 - \frac{1}{2}(1-p) = \frac{1 + p}{2}$.

\paragraph{Noise-based anonymization.}
Noise-based anonymization techniques are a class of methods that aim to protect the privacy of individuals in a dataset by adding random noise to the data. These can be seen as synthetic data generators, as they produce a synthetic dataset (of the same size as the original) given the original dataset.
In this work, we used the \texttt{aindo-anonymize} library \citep{aindo-anonymize} which defines a noise-based anonymization (called perturbation) that acts independently on each feature of a row and is designed to preserve the marginal distributions of the original data. The level of noise is controlled by a parameter $\alpha \in [0,1]$, where $\alpha=0$ corresponds to no noise (i.e., the original data) and $\alpha=1$ corresponds to maximum noise (i.e., completely random data). By varying $\alpha$, we can control the amount of privacy risk in the synthetic dataset, with higher values of $\alpha$ corresponding to lower privacy risk.
If a feature $x$ is numerical, the applied stochastic transformation is
\begin{equation}
x' = f^{-1}\left(\sqrt{1-\alpha_{\text{num}}} \cdot f(x) + \sqrt{\alpha_{\text{num}}} \cdot \epsilon\right)
\end{equation}
where $\epsilon$ is a random value sampled from the standard normal distribution and $f$ is a quantile normalization function that maps the original data to a normal distribution (for the implementation we refer to the \texttt{sklearn} library).
If a feature $x$ is categorical, the stochastic transformation consists in sampling a new value from the marginal distribution of the feature in the original data with probability $\alpha_{\text{cat}}$, and keeping the original value with probability $1-\alpha_{\text{cat}}$.
In practice, in our experiments we used $\alpha_{\text{num}} = \alpha^3$ and $\alpha_{\text{cat}} = \alpha^2$, where $\alpha \in [0,1]$ is the noise level parameter.

\subsection{Quality Evaluation}
\label{app:quality_evaluation}

We evaluate the quality of the synthetic data generated by the SDGs by using a metric of fidelity (the similarity between the synthetic and real data) and a metric of utility (the performance of a downstream task on the synthetic data).
For fidelity, we use detection \citep{snoke2018general}, which is the AUROC score of a discriminator trained to distinguish between real and synthetic data. The lower the Detection score, the higher the fidelity of the synthetic data.
For utility, we use ML efficacy \citep{park2018data}, which is the score of a classifier trained on the synthetic data and evaluated on a held-out test set of real data. The score depends on the downstream task, which consists of a multi-class classification for the UK Census dataset, a regression for the California dataset and a binary classification for the Adult dataset. We then measure the utility score as the accuracy for the multi-class classification, the negative root mean square error for the regression and the AUROC score for the binary classification. The higher the ML efficacy score, the higher the utility of the synthetic data.
In our case, we measure the difference between the score of the classifier trained on synthetic data and the score of a classifier trained on real data, such that a score of 0 means that the synthetic data has the same utility as the real data, while a negative score means that the synthetic data has lower utility than the real data.

We implement both the discriminator for detection and the classifier/regressor for ML efficacy using XGBoost \citep{chen2015xgboost} with default settings, and we cross-validate it over 5 different splits of the data (80\% training, 20\% validation) and report the mean score on the validation set.
Moreover, we repeat the whole process for 4 different random seeds and report the mean and standard deviation of the scores across the seeds.
Both ML Efficacy and Detection take at most a few seconds to be computed.

\subsection{Datasets}

\label{app:datasets}
We used the following datasets for our experiments:
\begin{itemize}
  \item Adult dataset \citep{adult}: This dataset contains demographic information extracted from the 1994 US Census database. It is a standard benchmark in research on mixed-type tabular data.
  \item California housing dataset \citep{california}: Derived from the 1990 California census, this dataset provides information regarding housing prices and related features.
  \item UK Census dataset \citep{uk_census}: This dataset is an anonymized 1\% random sample of the 2011 Census Microdata Teaching File for England and Wales, published by the Office for National Statistics.
\end{itemize}
\begin{table}[h]
  \centering
  \caption{Dataset features.}
  \begin{tabular}{lccc}
    \toprule
    Dataset & \#Records & \#Numerical Columns & \#Categorical Columns \\
    \midrule
    Adult & 48,842 & 5 & 10 \\
    California & 20,640 & 9 & 0 \\
    UK Census & 569,741 & 0 & 17 \\
    \bottomrule
  \end{tabular}
  \label{tab:dataset_summary}
\end{table}

\section{Additional Results}
\label{app:additional_results}

\begin{figure}[!htbp]
  \centering
  \includegraphics[width=1.0\textwidth]{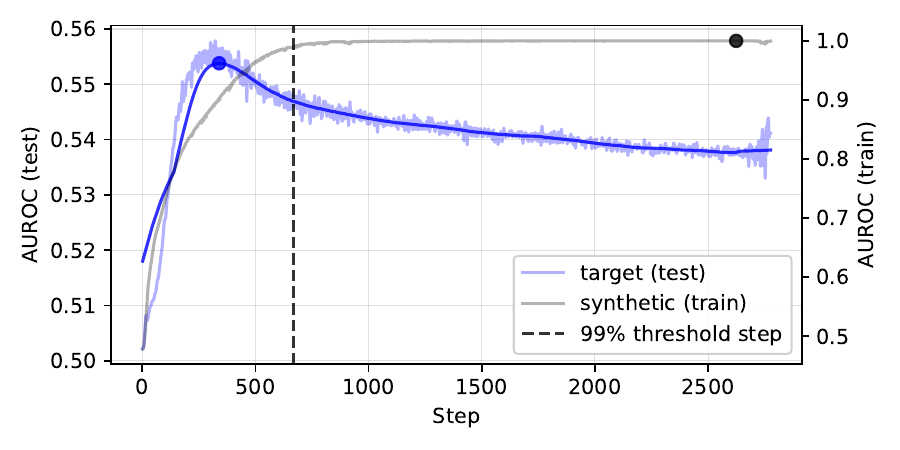}
  \caption{Example of ReMIA score as a function of learning step. We show the ReMIA score (AUROC) on the target records and on the synthetic data across learning steps for one of the runs of ADSGAN on UK Census dataset. The vertical dashed line represents the step when the performance on the target set is selected as the privacy score of ReMIA, which, according to the stopping criterion, is the step where the score on synthetic data reaches 99\%. In this example, the maximum value of the score on the target records (blue dot) is reached earlier than the step indicated by the stopping criterion leading to the privacy score and the maximum value on the synthetic data (black dot).
  }
  \label{remia_score_learning_step}
\end{figure}

\begin{figure}[!htbp]
  \centering
  \includegraphics[width=1.0\textwidth]{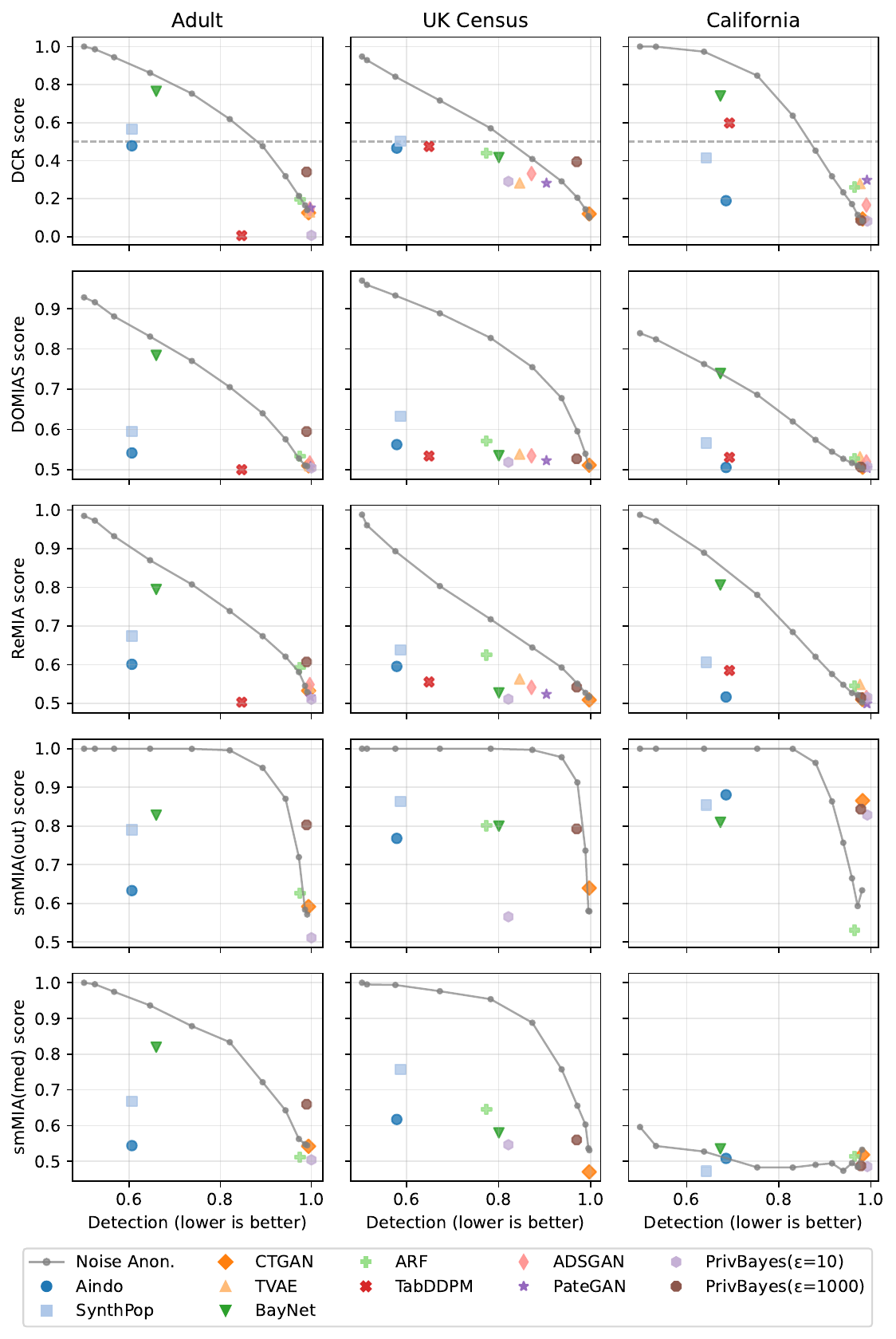}
  \caption{Privacy-fidelity trade-off. We show the fidelity of the synthetic data in terms of Detection (the lower the better) against the privacy scores (the lower the better) of the different privacy metrics for all three datasets. The gray line represents the performance of the noise-based anonymization method.
  }
  \label{fidelity_vs_privacy}
\end{figure}

\begin{figure}[!htbp]
  \centering
  \includegraphics[width=1.0\textwidth]{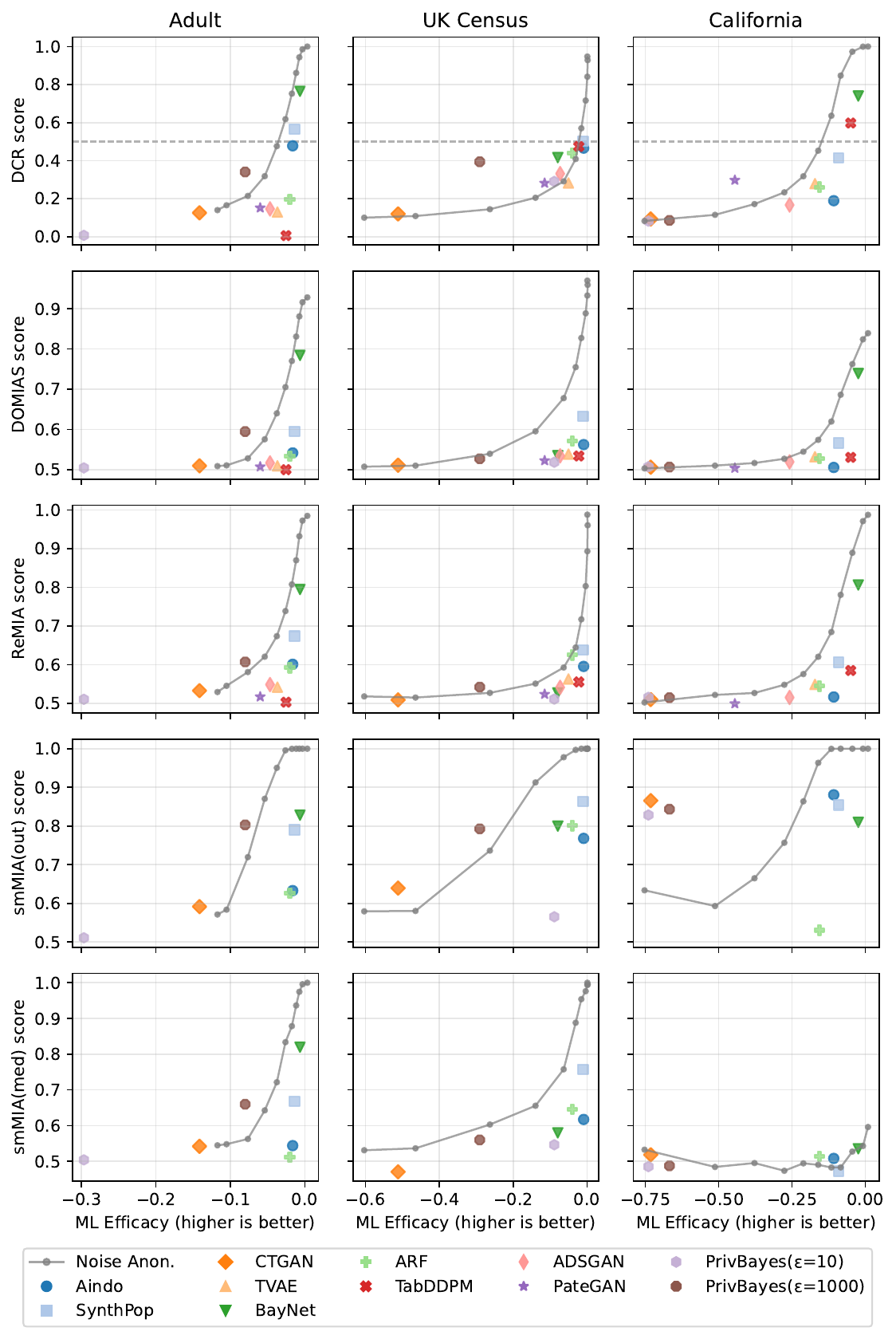}
  \caption{Privacy-utility trade-off. We show the utility of the synthetic data in terms of ML Efficacy (the higher the better) against the privacy scores (the lower the better) of the different privacy metrics for all three datasets. The gray line represents the performance of the noise-based anonymization method.
  }
  \label{utility_vs_privacy}
\end{figure}

\begin{figure}[!htbp]
  \centering
  \includegraphics[width=1.0\textwidth]{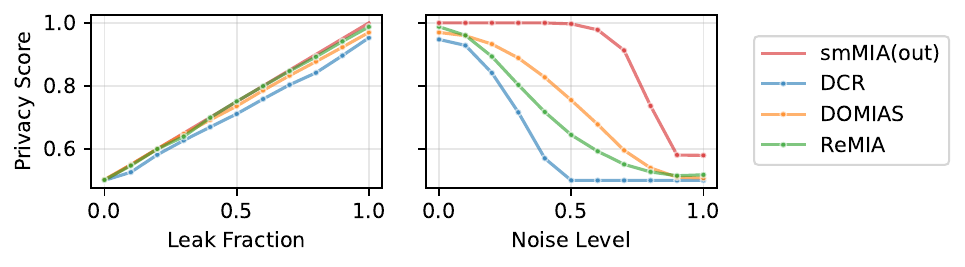}
  \caption{Metrics evaluation in SDGs with controlled level of privacy risk (UK Census dataset).
  }
  \label{leak_fraction_alpha_vs_privacy_uk_census}
\end{figure}

\begin{figure}[!htbp]
  \centering
  \includegraphics[width=1.0\textwidth]{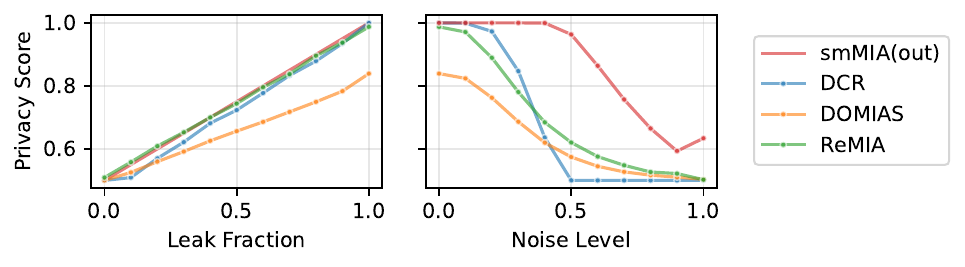}
  \caption{Metrics evaluation in SDGs with controlled level of privacy risk (California dataset).
  }
  \label{leak_fraction_alpha_vs_privacy_california}
\end{figure}

This section collects supplementary experimental results and figures referenced in the main text.
Figure \ref{remia_score_learning_step} shows how the AUROC on synthetic (train) and target (test) data in ReMIA evolves across learning steps.
Table \ref{tab:privacy_scores_uk_census} reports privacy scores on UK Census, Table \ref{tab:privacy_scores_california} reports privacy scores on California, and Tables \ref{tab:detection} and \ref{tab:ml_efficacy} summarize fidelity and utility across datasets and SDGs. Figure \ref{leak_fraction_alpha_vs_privacy_uk_census} shows privacy risk under controlled leak and noise on UK Census, Figure \ref{leak_fraction_alpha_vs_privacy_california} shows the same analysis on California, and Figures \ref{fidelity_vs_privacy} and \ref{utility_vs_privacy} present the privacy-quality trade-off for all privacy metrics and datasets.

\begin{table}
  \caption{Privacy scores (AUROC) for UK Census dataset (\%). We report mean and standard deviation over 4 different seeds.}
  \label{tab:privacy_scores_uk_census}
  \centering
  \begin{tabular}{lS[table-format=2.1(1)]S[table-format=2.1(1)]S[table-format=2.1(1)]S[table-format=2.1(1)]S[table-format=2.1(1)]}
    \toprule
    Generator & \multicolumn{1}{c}{DCR} & \multicolumn{1}{c}{DOMIAS} & \multicolumn{1}{c}{ReMIA} & \multicolumn{1}{c}{smMIA(out)} & \multicolumn{1}{c}{smMIA(med)} \\
    \midrule
    ADSGAN & \num{33.1 +- 3.9} & \underline{\num{53.4 +- 1.2}} & \underline{\num{54.1 +- 0.7}} & \multicolumn{1}{c}{-} & \multicolumn{1}{c}{-} \\
    ARF & \num{43.9 +- 1.0} & \underline{\num{57.2 +- 1.5}} & \underline{\num{62.5 +- 1.8}} & \underline{\num{80.1 +- 2.1}} & \underline{\num{64.5 +- 6.4}} \\
    Aindo & \num{46.6 +- 1.9} & \underline{\num{56.2 +- 0.7}} & \underline{\num{59.5 +- 0.4}} & \underline{\num{76.8}} & \underline{\num{61.7}} \\
    BayNet & \num{41.6 +- 3.5} & \underline{\num{53.5 +- 2.0}} & \underline{\num{52.7 +- 1.3}} & \underline{\num{80.0 +- 5.4}} & \num{58.0 +- 2.7} \\
    CTGAN & \num{11.9 +- 2.7} & \num{51.1 +- 0.9} & \num{50.9 +- 1.3} & \underline{\num{64.0 +- 1.7}} & \num{47.1 +- 2.8} \\
    PateGAN & \num{28.1 +- 1.5} & \num{52.3 +- 1.1} & \num{52.4 +- 0.2} & \multicolumn{1}{c}{-} & \multicolumn{1}{c}{-} \\
    PrivBayes($\epsilon$=10) & \num{29.0 +- 4.9} & \underline{\num{51.8 +- 1.3}} & \num{51.1 +- 1.2} & \num{56.5 +- 0.8} & \num{54.6 +- 2.9} \\
    PrivBayes($\epsilon$=1000) & \num{39.4 +- 3.2} & \underline{\num{52.7 +- 1.0}} & \underline{\num{54.2 +- 1.7}} & \underline{\num{79.3 +- 2.0}} & \num{56.0 +- 2.2} \\
    SynthPop & \num{50.2 +- 2.8} & \underline{\num{63.3 +- 1.0}} & \underline{\num{63.8 +- 1.2}} & \underline{\num{86.4 +- 2.3}} & \underline{\num{75.7 +- 2.1}} \\
    TVAE & \num{28.2 +- 8.1} & \underline{\num{53.9 +- 0.9}} & \underline{\num{56.3 +- 1.5}} & \multicolumn{1}{c}{-} & \multicolumn{1}{c}{-} \\
    TabDDPM & \num{47.4 +- 18.9} & \underline{\num{53.4 +- 2.3}} & \underline{\num{55.5 +- 1.9}} & \multicolumn{1}{c}{-} & \multicolumn{1}{c}{-} \\
    \bottomrule
\end{tabular}

\end{table}

\begin{table}
  \caption{Privacy scores (AUROC) for California dataset (\%). We report mean and standard deviation over 4 different seeds.}
  \label{tab:privacy_scores_california}
  \centering
  \begin{tabular}{lS[table-format=2.1(1)]S[table-format=2.1(1)]S[table-format=2.1(1)]S[table-format=2.1(1)]S[table-format=2.1(1)]}
    \toprule
    Generator & \multicolumn{1}{c}{DCR} & \multicolumn{1}{c}{DOMIAS} & \multicolumn{1}{c}{ReMIA} & \multicolumn{1}{c}{smMIA(out)} & \multicolumn{1}{c}{smMIA(med)} \\
    \midrule
    ADSGAN & \num{16.6 +- 4.7} & \num{51.9 +- 1.6} & \num{51.5 +- 1.6} & \multicolumn{1}{c}{-} & \multicolumn{1}{c}{-} \\
    ARF & \num{26.0 +- 6.5} & \underline{\num{52.8 +- 0.8}} & \underline{\num{54.4 +- 1.9}} & \num{53.1 +- 5.7} & \num{51.4 +- 4.0} \\
    Aindo & \num{18.9 +- 2.2} & \num{50.6 +- 0.4} & \num{51.6 +- 1.6} & \underline{\num{88.1}} & \num{50.8} \\
    BayNet & \underline{\num{74.0 +- 4.6}} & \underline{\num{73.9 +- 2.0}} & \underline{\num{80.6 +- 0.9}} & \underline{\num{81.0 +- 6.1}} & \num{53.5 +- 5.7} \\
    CTGAN & \num{9.3 +- 3.2} & \num{50.6 +- 0.7} & \num{51.0 +- 1.6} & \underline{\num{86.6 +- 4.2}} & \num{51.8 +- 3.6} \\
    PateGAN & \num{29.7 +- 16.4} & \num{50.4 +- 1.1} & \num{49.9 +- 1.1} & \multicolumn{1}{c}{-} & \multicolumn{1}{c}{-} \\
    PrivBayes($\epsilon$=10) & \num{8.1 +- 2.2} & \num{50.7 +- 0.9} & \num{51.6 +- 1.3} & \underline{\num{82.9 +- 1.9}} & \num{48.5 +- 2.5} \\
    PrivBayes($\epsilon$=1000) & \num{8.5 +- 2.4} & \num{50.7 +- 1.2} & \num{51.4 +- 0.9} & \underline{\num{84.3 +- 4.6}} & \num{48.8 +- 2.2} \\
    SynthPop & \num{41.4 +- 9.1} & \underline{\num{56.6 +- 1.3}} & \underline{\num{60.6 +- 1.1}} & \underline{\num{85.5 +- 2.4}} & \num{47.2 +- 3.1} \\
    TVAE & \num{27.9 +- 3.4} & \underline{\num{53.2 +- 1.1}} & \underline{\num{54.9 +- 1.4}} & \multicolumn{1}{c}{-} & \multicolumn{1}{c}{-} \\
    TabDDPM & \underline{\num{59.8 +- 17.2}} & \underline{\num{53.1 +- 1.5}} & \underline{\num{58.5 +- 0.7}} & \multicolumn{1}{c}{-} & \multicolumn{1}{c}{-} \\
    \bottomrule
\end{tabular}

\end{table}

\begin{table}
  \caption{Detection scores  (\%). This table reports the Detection (the lower the better) scores for all SDGs and datasets. SDGs were evaluated on a subset of 10,000 records from the original datasets, with the exception of CTGAN where we used 5,000 records. We report mean and standard deviation over 4 different seeds. See Section \ref{app:experimental_setup} for more details on how detection is computed.}
  \label{tab:detection}
  \centering
  \begin{tabular}{lS[table-format=3.2(2)]S[table-format=2.2(2)]S[table-format=2.2(2)]}
    \toprule
    Generator & \multicolumn{1}{c}{Adult} & \multicolumn{1}{c}{California} & \multicolumn{1}{c}{UK Census} \\
    \midrule
    ADSGAN & \num{99.65 +- 0.26} & \num{99.05 +- 0.26} & \num{87.12 +- 2.26} \\
    ARF & \num{97.50 +- 0.22} & \num{96.52 +- 0.10} & \num{77.35 +- 1.02} \\
    Aindo & \num{60.58 +- 1.04} & \num{68.55 +- 1.18} & \num{57.82 +- 0.40} \\
    BayNet & \num{65.92 +- 0.82} & \num{67.33 +- 0.31} & \num{80.03 +- 0.66} \\
    CTGAN & \num{99.38 +- 0.13} & \num{98.22 +- 0.17} & \num{99.67 +- 0.13} \\
    PateGAN & \num{99.80 +- 0.14} & \num{99.17 +- 0.21} & \num{90.38 +- 1.16} \\
    PrivBayes($\epsilon$=10) & \num{100.00 +- 0.00} & \num{99.25 +- 0.06} & \num{82.10 +- 2.04} \\
    PrivBayes($\epsilon$=1000) & \num{98.95 +- 0.24} & \num{97.85 +- 1.08} & \num{96.95 +- 1.14} \\
    SynthPop & \num{60.60 +- 0.36} & \num{64.30 +- 0.57} & \num{58.67 +- 1.01} \\
    TVAE & \num{99.60 +- 0.12} & \num{97.65 +- 0.17} & \num{84.55 +- 1.34} \\
    TabDDPM & \num{84.70 +- 0.42} & \num{69.30 +- 0.67} & \num{64.85 +- 0.62} \\
    \bottomrule
\end{tabular}

\end{table}

\begin{table}
  \caption{ML efficacy scores (\%). This table reports the ML Efficacy (the higher the better) scores for all SDGs and datasets. SDGs were evaluated on a subset of 10,000 records from the original datasets, with the exception of CTGAN where we used 5,000 records. We report mean and standard deviation over 4 different seeds. See Section \ref{app:experimental_setup} for more details on how ML efficacy is computed.}
  \label{tab:ml_efficacy}
  \centering
  \begin{tabular}{lS[table-format=2.2(2)]S[table-format=2.2(2)]S[table-format=2.2(2)]}
    \toprule
    Generator & \multicolumn{1}{c}{Adult} & \multicolumn{1}{c}{California} & \multicolumn{1}{c}{UK Census} \\
    \midrule
    ADSGAN & \num{-4.68 +- 1.00} & \num{-25.79 +- 3.49} & \num{-7.28 +- 0.44} \\
    ARF & \num{-2.02 +- 0.73} & \num{-15.56 +- 0.59} & \num{-4.01 +- 0.70} \\
    Aindo & \num{-1.63 +- 0.61} & \num{-10.74 +- 1.64} & \num{-0.91 +- 0.29} \\
    BayNet & \num{-0.65 +- 0.51} & \num{-2.37 +- 0.49} & \num{-7.92 +- 0.39} \\
    CTGAN & \num{-14.14 +- 2.28} & \num{-73.14 +- 1.75} & \num{-51.18 +- 3.26} \\
    PateGAN & \num{-5.99 +- 0.79} & \num{-44.45 +- 9.06} & \num{-11.48 +- 0.76} \\
    PrivBayes($\epsilon$=10) & \num{-29.68 +- 13.25} & \num{-73.93 +- 2.69} & \num{-8.87 +- 2.55} \\
    PrivBayes($\epsilon$=1000) & \num{-8.02 +- 3.26} & \num{-66.76 +- 3.57} & \num{-29.05 +- 19.54} \\
    SynthPop & \num{-1.38 +- 0.38} & \num{-9.12 +- 1.12} & \num{-1.08 +- 0.28} \\
    TVAE & \num{-3.68 +- 0.50} & \num{-17.13 +- 2.22} & \num{-5.03 +- 0.72} \\
    TabDDPM & \num{-2.52 +- 0.50} & \num{-4.96 +- 0.67} & \num{-2.22 +- 0.17} \\
    \bottomrule
\end{tabular}

\end{table}

\section{Resources and Assets}

\subsection{Computational Resources}
\label{app:computational_resources}

The following hardware was used for performing the experiments in this article:
\begin{itemize}
    \item CPU: AMD Ryzen Threadripper 2950X (16-Core)
    \item Memory: 125 GiB
    \item GPU: NVIDIA RTX A5000
\end{itemize}

Running the experiments for the paper took approximately 882 hours of computational time (assuming serial execution), of which about 865 hours were spent on attacks based on shadow modeling.
We estimate that the total time taken for all experiments (including those not reported in the paper) was approximately 1500 hours.

\subsection{Assets and Code Availability}
\label{app:assets_code_availability}
Our code will be released upon publication, excluding the proprietary generative model used in our experiments.
All the assets and code used in our experiments, with the exception of the proprietary generative model, are open-source and publicly available.
We summarize the relevant assets used in our experiments and their availability in Table \ref{tab:assets_summary}.
\begin{table}[h]
  \centering
  \caption{Summary of assets used in our experiments.}
  \begin{tabular}{lcc}
    \toprule
    Asset & License & Credits \\
    \midrule
    Adult dataset & CC BY 4.0 & \citet{adult} \\
    California dataset & CC0 public domain & \citet{california} \\
    UK Census dataset & Open Government Licence (OGL) & \citet{uk_census} \\
    aindo-anonymize library & MIT License & \citet{aindo-anonymize} \\
    DOMIAS code & MIT License & \citet{van2023membership} \\
    smMIA code & MIT License & \citet{meeus2023achilles} \\
    \bottomrule
  \end{tabular}
  \label{tab:assets_summary}
\end{table}

\end{document}